\documentclass[sigconf]{acmart}
\usepackage{algorithm}
\usepackage{algorithmic}
\usepackage{amsmath}
\usepackage{booktabs}
\usepackage{graphicx}
\usepackage{multirow}
\usepackage{balance}
\usepackage{enumitem}
\floatstyle{ruled}\restylefloat{algorithm}
\usepackage[utf8]{inputenc}
\usepackage[export]{adjustbox}

\setcopyright{rightsretained}
\copyrightyear{2025}\acmYear{2025}
\acmConference[KDD-UMC '25]{KDD 2025 Undergraduate and Master’s Consortium}{August 03-07, 2025}{Toronto, Canada}

\begin{document}

\title[SIFOTL]{SIFOTL: A Principled, Statistically-Informed Fidelity-Optimization Method for Tabular Learning}

\author{Shubham Mohole}
\email{sam588@cornell.edu}
\affiliation{%
  \institution{Cornell University}
  \city{Ithaca}
  \state{New York}
  \country{USA}
}

\author{Sainyam Galhotra}
\email{sg@cs.cornell.edu}
\affiliation{%
  \institution{Cornell University}
  \city{Ithaca}
  \state{New York}
  \country{USA}
}

\begin{abstract}
Identifying the factors driving data shifts in tabular datasets is a significant challenge for analysis and decision support systems, especially those focusing on healthcare. Privacy rules restrict data access, and noise from complex processes hinders analysis. To address this challenge, we propose SIFOTL (Statistically-Informed Fidelity-Optimization Method for Tabular Learning) that (i) extracts privacy-compliant data summary statistics, (ii) employs twin XGBoost models to disentangle intervention signals from noise with assistance from LLMs, and (iii) merges XGBoost outputs via a Pareto-weighted decision tree to identify interpretable segments responsible for the shift. Unlike existing analyses which may ignore noise or require full data access for LLM-based analysis, SIFOTL addresses both challenges using only privacy-safe summary statistics. Demonstrating its real-world efficacy, for a MEPS panel dataset mimicking a new Medicare drug subsidy, SIFOTL achieves an F1 score of 0.85, substantially outperforming BigQuery Contribution Analysis (F1=0.46) and statistical tests (F1=0.20) in identifying the segment receiving the subsidy. Furthermore, across 18 diverse EHR datasets generated based on Synthea ABM, SIFOTL sustains F1 scores of 0.86–0.96 without noise and $\geq$ 0.75 even with injected observational noise, whereas baseline average F1 scores range from 0.19–0.67 under the same tests. SIFOTL, therefore, provides an interpretable, privacy‑conscious workflow that is empirically robust to observational noise.
\end{abstract}



\maketitle

\section{Introduction}
\begin{figure}
\centering
\begin{adjustbox}{max width=\columnwidth}
  \includegraphics{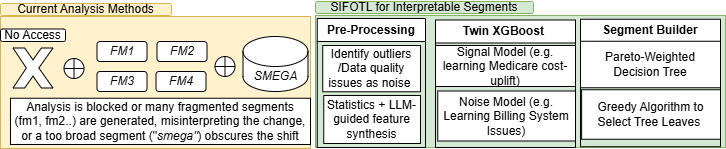}
  
\end{adjustbox}
\caption{Illustrative comparison in the context of our experiments - current healthcare data analysis (left) versus the SIFOTL method (right).}
\label{fig:inro}
\end{figure}

Timely detection of distribution shifts is increasingly important for evidence‑based policy and clinical decision-making. This analytical challenge is particularly salient in healthcare and public policy domains, where data is often subject to stringent privacy regulations and contains observational noise. At the same time, such analysis-driven decisions have immense consequences.

As Figure \ref{fig:inro} highlights, although analysis is often mission-critical, first, data access restrictions severely limit analysts' ability to perform comprehensive contribution analysis. As highlighted by Wartenberg and Thompson \cite{wartenberg2010privacy}, health researchers face mounting barriers to accessing vital data, with laws like HIPAA and the Family Educational Rights and Privacy Act setting specific data compliance rules. In some cases, these restrictions have even hindered analyses that could identify key drivers of health outcomes. Native American public health officials reported being “\emph{blinded}” during critical public health emergencies, with tribal epidemiology centers often denied access to data available to other public health workers despite legal requirements for data sharing \cite{rodriguez2024native}. During the COVID-19 pandemic, this problem became especially pronounced, with tribal health officials documenting that data denials "impeded their responses to disease outbreaks" \cite{rodriguez2024native}.

Second, even when data access is granted, data quality issues threaten the reliability of contribution analyses. Yang et al. \cite{yang2024addressing} demonstrate that electronic health records (EHRs) are governed by numerous regulatory constraints that limit access, while simultaneously containing significant noise from "data entry errors, incomplete information, inconsistencies, system errors, and diagnostic test errors." Third, the quality of the analysis itself leaves much to be desired. Finlayson et al. \cite{finlayson2021clinician} document how analytical systems in healthcare settings frequently lead to incorrect conclusions due to "dataset shift"—where patterns identified during development fail to generalize during deployment. At the University of Michigan Hospital, a sepsis-alerting model had to be deactivated during the COVID-19 pandemic due to "spurious alerting owing to changes in patients' demographic characteristics" \cite{finlayson2021clinician}, highlighting how undetected changes in population characteristics can compromise automated systems if the data shifts are not properly monitored and identified.

Although identifying contribution factors is of great significance, the traditional methods used for this analysis task often fall short. Contribution analysis techniques like BigQuery tend to emphasize aggregate changes that may overlook important subpopulation effects. Statistical approaches face challenges in distinguishing intervention signals from noise artifacts, especially when data quality varies across sources or time periods and requires domain-level knowledge. Moreover, privacy-preserving methods may sacrifice analytical power in the name of confidentiality.

SIFOTL addresses these challenges through a privacy-preserving architecture that disentangles real signals from noise while providing interpretable insights without exposing sensitive data. Our research examines how analysts can (1) identify contributing population segments amid varying noise levels, (2) leverage privacy-compliant LLM-guided feature synthesis using only statistical summaries, and (3) maintain robustness across diverse intervention types and noise conditions.

Our contributions include a two-stage analytical framework employing twin XGBoost models to identify signals and noise separately, a privacy-preserving approach where LLMs operate solely on statistical summaries, a Pareto-weighted decision tree optimization balancing signal coverage with noise exclusion, and empirical evaluation results showing improved accuracy in identifying real contribution factors.

\section{SIFOTL Architecture}
The SIFOTL method, through its coordinated components, systematically identifies true intervention effects while discounting noise.

\subsection{System Overview}
Operating on control and test datasets, SIFOTL follows this sequence of steps:
\begin{itemize}
    \item \textbf{Statistical Pre-analysis}: A set of statistical tests ($\chi^2$ test, Point-Biserial Correlation, Cramer's V effect size) on both raw data and numerical data binned into categories generate privacy-safe summary insights comparing the test and control datasets, separately focusing on potential intervention signals (comparing groups based on a target metric difference) and potential noise patterns. We have made the results of these tests available in the GitHub repository for the data in our experiments.
    \item \textbf{Noise-Inference Preprocessing}: Rule-based heuristics identify noise patterns (e.g., duplicates). In the GitHub repository, we share the overview of these rules, but these rules are configurable in the system and can be extended as per the domain user's requirements.
    \item \textbf{LLM-Guided Feature Synthesis}: A privacy-preserving interaction where a Large Language Model (LLM) receives only the statistical summaries (not raw data) and generates candidate feature definitions and corresponding Python code tailored for predicting either the intervention signal (Model C) or the noise pattern (Model N). (Details in Section \ref{sec:llm_feature_synthesis}).
    \item \textbf{Twin XGBoost  Models}: An intervention predictor (Model C) trained on the target metric difference using its LLM-generated features, and a noise predictor (Model N) trained on noise labels using its distinct LLM-generated features. These produce row-level probabilities $p_C$ and $p_N$. (Details in Section \ref{sec:prob_tree_pipeline}).
    \item \textbf{Pareto-Weighted Decision Tree}: An interpretable tree trained on the intervention signal, using the outputs ($p_C, p_N$) to adaptively weight samples, balancing signal fidelity and noise exclusion by optimizing a penalty parameter $\alpha$. (Details in Section \ref{sec:prob_tree_pipeline}).
    \item \textbf{Segment Extraction}: A final segment definition is extracted from the optimized tree based on user-defined criteria (e.g., mass threshold). (Details in Section \ref{sec:prob_tree_pipeline}).
\end{itemize}

\subsection{LLM-Guided Feature Synthesis}\label{sec:llm_feature_synthesis}
One of the important features of the SIFOTL method is its privacy-preserving feature engineering process, which leverages an LLM operating solely on statistical summaries derived from the data, rather than the raw data itself. This process generates tailored features for both the intervention prediction task (Model C) and the noise prediction task (Model N).

The overall workflow involves two main LLM interactions per model (C and N):
\begin{enumerate}
    \item \textbf{Feature Definition Generation:} The LLM receives statistical insights (e.g., significant differences between target groups based on $\chi^2$ test, correlations) and the base data schema (excluding sensitive or target columns). Before any large-language-model (LLM) prompts are issued, we enforce \emph{dataset-level} and \emph{slice-level} safeguards to guarantee that no intermediate view can single out an individual. During association testing, our pipeline iterates over every category (or numeric bin) of each contributor feature, forming a \emph{slice}—the set of rows that satisfy that single condition. Before a slice is admitted for hypothesis testing or index export, we run an anonymity check: the slice must contain at least \texttt{MIN\_ANON\_SLICE\_SIZE} rows, and—when quasi-identifier columns are available—those rows must also satisfy \(k\!\ge\! \texttt{K\_ANON\_THRESHOLD}\) anonymity as verified by \texttt{pycanon}.  Slices that fail either criterion are suppressed. These parameters are adaptive based on the dataset size (with a minimum value of two), and they can be tightened without changing the surrounding pipeline logic. With such insights and dataset schema, LLM is prompted to propose new engineered features relevant to the specific prediction task (intervention or noise), describing their logic and potential value based on the provided statistics.
    \item \textbf{Code Generation:} The LLM receives the validated feature definitions (including names, source columns, and logic descriptions) generated in the previous step and is prompted to write a Python/Pandas function that implements the calculation logic for these features.
\end{enumerate}

Algorithm 1 outlines this process. Note that this algorithm is executed separately for Model C and Model N.

\begin{algorithm}[H]
  \caption{\textsc{LLMFeatureSynthesis}}
  \label{alg:llm_synthesis}
  \begin{algorithmic}[1]
    \STATE \textbf{Input:}\; statistical insights $I$; base schema $S$;
           excluded cols (target, ground truth labels).\ $E$; LLM endpoint $L_{api}$
    \STATE \textbf{Output:}\; feature list $\mathcal{F}_{\text{def}}$,
           code path $P_{code}$\\[2pt]
    \STATE \COMMENT{\textit{Feature-definition phase}}
    \STATE $S_{\text{context}} \leftarrow S \setminus E$
    \STATE $I_{\text{summary}} \leftarrow \textsc{SummarizeInsights}(I)$
    \STATE $\textit{prompt}_{\text{def}} \leftarrow
           \textsc{ConstrDefinitionPrompt}(S_{\text{context}}, I_{\text{summary}})$
    \STATE $\textit{response}_{\text{def}} \leftarrow
           \textsc{CallLLM}(L_{api},\textit{prompt}_{\text{def}})$
    \STATE $\mathcal{F}_{\text{def}} \leftarrow
           \textsc{ParseValidateDefs}(\textit{response}_{\text{def}})$
    \IF{$\mathcal{F}_{\text{def}} = \varnothing$}
      \STATE \textbf{return} \textit{Error}
    \ENDIF
    \STATE \textsc{StoreDefs}$(\mathcal{F}_{\text{def}})$\\[2pt]
    \STATE \COMMENT{\textit{Code-generation phase}}
    \STATE $\textit{prompt}_{\text{code}} \leftarrow
           \textsc{ConstrCodePrompt}(\mathcal{F}_{\text{def}})$
    \STATE $\textit{response}_{\text{code}} \leftarrow
           \textsc{CallLLM}(L_{api},\textit{prompt}_{\text{code}})$
    \STATE $\textit{code\_content} \leftarrow
           \textsc{ExtractPyCode}(\textit{response}_{\text{code}})$
    \IF{$\textit{code\_content}$ \textbf{invalid}}
      \STATE \textbf{return} \textit{Error}
    \ENDIF
    \STATE $P_{code} \leftarrow
           \textsc{SaveCode}(\textit{code\_content},\mathcal{F}_{\text{def}})$
    \STATE \textsc{UpdateDefStatus}$(\mathcal{F}_{\text{def}},P_{code},
           \text{'code\_gen'})$
    \STATE \textbf{return} $\mathcal{F}_{\text{def}}, P_{code}$
  \end{algorithmic}
\end{algorithm}

The generated code file ($P_{code}$) contains a Python function that reads and then populates feature values in the database, making them available for the XGBoost models. Currently, any failures are processed offline, and automating the detection and correction of these failures remains the focus of our future work.

\subsection{Step-by-Step SIFOTL Algorithm: Probabilistic Labeling and Tree Search}\label{sec:prob_tree_pipeline}

This section details the \emph{probabilistic-plus-tree} pipeline, which uses the features generated via LLM synthesis (Section \ref{sec:llm_feature_synthesis}) and the noise inference preprocessing.

\begin{enumerate}[leftmargin=*, label=\textbf{\arabic*.}]
    \item \textbf{Probabilistic labeling (Stage 1).} \label{step:stage1}
          Two disjoint XGBoost models (Model C for intervention, Model N for noise), trained on their respective LLM-generated feature sets, produce row-level probabilities
          \[
              f_{\!C}:X_C\!\rightarrow\!p_C,
              \qquad
              f_{\!N}:X_N\!\rightarrow\!p_N 
          \]
          estimating, respectively, intervention membership ($p_C$) and observational noise ($p_N$).

    \item \textbf{Weighted decision-tree search (Stage 2).} \label{step:stage2}
          For every candidate penalty $\alpha$ in a predefined grid $\mathcal{A}$, compute the per-row weight:
          \[ \label{eq:weight_def_alt} 
              w_i(\alpha)=
              \frac{p_C(i)}{p_C(i)+\alpha\,p_N(i)+\varepsilon},
              \qquad
              \varepsilon=10^{-9},
          \]
          Fit a shallow decision tree $T_\alpha$ \cite{quinlan1986induction} on features $X$
          using the \emph{publicly observable} metric-difference indicator
          $\tilde y_i$ as the class label and $w_i(\alpha)$ as
          \texttt{sample\_weight}.
          Each $T_\alpha$ is scored in the $(M_{\text{signal}},M_{\text{noise}})$
plane (definitions below), a common approach in multi-objective optimization problems (e.g., \cite{horn2015model, martinez2020minimax}); the $\alpha^\star$ at the empirical Pareto "knee" is selected.

    \item \textbf{Greedy segment extraction (Stage 3).} \label{step:stage3}
          Starting from the optimal tree $T_{\alpha^\star}$, iteratively add the leaves predicted as class 1, sorted by purity (e.g., average $p_C$), until the cumulative $p_C$ mass of the segment reaches a user-defined threshold $\tau$.
\end{enumerate}

\vspace{0.5em}
\textbf{Per-row primitives}
\begin{itemize}[leftmargin=*, itemsep=0.25em]
    \item $p_C(i)$ — probability of belonging to the intervention slice (from Model C).
    \item $p_N(i)$ — probability of being noisy (from Model N).
    \item $w_i(\alpha)$ — adaptive training weight from Eq.\,\eqref{eq:weight_def_alt}. 
    \item $\tilde y_i$ — \emph{surrogate} binary label (e.g., indicator of target metric difference); no hidden ground truth used here.
\end{itemize}

\textbf{Tree-level objectives (for Pareto optimization)}
\[
\begin{aligned}
M_{\text{signal}}(T) &= \sum_{\ell \in \text{leaves}(T)} \left( \sum_{i \in \text{leaf}_\ell} w_i(\alpha) \right) \times \overline{p_C}^{\,\ell}, \\
& \quad \text{(Weighted \(p_C\) mass via leaf averages)}\\[3pt]
M_{\text{noise}}(T)  &= 1-\bigl\lvert\operatorname{corr}_{\ell}\!\bigl(\overline{p_C}^{\,\ell},\overline{p_N}^{\,\ell}\bigr)\bigr\rvert, \\
& \quad \text{(Noise robustness via correlation - larger the better)}
\end{aligned}
\]

where the correlation and averages ($\overline{p_\bullet}^{\,\ell}$) are taken across leaves $\ell$ of the tree $T_\alpha$. $\overline{p_C}^{\,\ell}$ and $\overline{p_N}^{\,\ell}$ are the weighted averages of $p_C$ and $p_N$ for rows within leaf $\ell$, respectively, using weights $w_i(\alpha)$.
\subsubsection*{Algorithm 2: Pareto-Weighted Tree Search}
\begin{algorithm}[H]
  \caption{\textsc{WeightedTreeSearch}}
  \label{alg:pareto}
  \begin{algorithmic}[1]
    \REQUIRE feature matrix $X$; surrogate label $\tilde y$;
             probabilities $(p_C, p_N)$
    \ENSURE  optimal tree $T^\star$, optimal penalty $\alpha^\star$
    \STATE choose grid $\mathcal{A} = \{\alpha_{\min},\dots,\alpha_{\max}\}$
    \FOR{$\alpha \in \mathcal{A}$}
      \STATE $w \leftarrow w(\alpha)$  \COMMENT{see Eq.\,\eqref{eq:weight_def_alt}}
      \STATE $T_\alpha \leftarrow \textsc{FitTree}(X,\tilde y,w)$
             \COMMENT{fit shallow decision tree}
      \STATE evaluate $M_{\text{signal}}(T_\alpha)$ and
             $M_{\text{noise}}(T_\alpha)$
    \ENDFOR
    \STATE $(\textit{score}_{\text{sig}},\textit{score}_{\text{noise}})
           \leftarrow$ all metrics across $\mathcal{A}$
    \STATE $\alpha^\star \leftarrow
           \textsc{KneePoint}(\textit{score}_{\text{sig}},
           \textit{score}_{\text{noise}})$
    \STATE $T^\star \leftarrow T_{\alpha^\star}$
    \RETURN $T^\star, \alpha^\star$
  \end{algorithmic}
\end{algorithm}
\subsubsection*{Algorithm 3: Mass-Greedy Segment Selection}
\begin{algorithm}[H]
  \caption{\textsc{MassGreedy}}
  \label{alg:greedy}
  \begin{algorithmic}[1]
    \REQUIRE leaves of $T^\star$; probability vector $p_C$; mass threshold $\tau$
    \ENSURE  binary mask $\sigma$ of the final segment
    \STATE $\mathcal{L} \leftarrow$ leaves in $T^\star$ with predicted class 1
    \STATE sort $\mathcal{L}$ by descending mean $p_C$
    \STATE $\sigma \leftarrow \mathbf{0}$  \COMMENT{size = num\_rows}
    \STATE $m_{\text{curr}} \leftarrow 0$;\;
           $m_{\text{tgt}} \leftarrow \tau \sum_i p_C(i)$
    \FOR{leaf $\ell$ in $\mathcal{L}$}
      \IF{$m_{\text{curr}} \ge m_{\text{tgt}}$}
        \STATE \textbf{break}
      \ENDIF
      \STATE $\textit{mask}_\ell \leftarrow
             \textsc{getRowsInLeaf}(\ell)$
      \STATE $\sigma \leftarrow \sigma \lor \textit{mask}_\ell$
      \STATE $m_{\text{curr}} \leftarrow
             \sum_{i:\sigma_i = 1} p_C(i)$
    \ENDFOR
    \RETURN $\sigma$
  \end{algorithmic}
\end{algorithm}

\paragraph{Interpretation.}
The tree captures regions that are consistently high in intervention probability $p_C$ while penalizing co-occurrence with high noise probability $p_N$ via the $\alpha$-weighted samples. The greedy post-processing selects the most confident leaves (in terms of $p_C$) to construct a high-fidelity segment meeting the desired mass coverage $\tau$. Crucially, this segmentation relies only on the model outputs ($p_C$, $p_N$) and the surrogate label $\tilde y$, not hidden ground-truth intervention flags.

\paragraph{Why Fractional Weights Instead of Hard Filtering?} Using $w_i(\alpha)$ from Eq.\,\eqref{eq:weight_def_alt} allows the tree search (Algorithm 2) to \emph{observe} noisy rows—thereby preserving geometric context within the feature space—yet discourages split decisions dominated by high-$p_N$ regions. This soft-weighting strategy, implemented via \texttt{sample\_weight} in the tree fitting process (along with optional \texttt{class\_weight} from config, e.g., \texttt{\{0:1, 1:10\}}), empirically yields higher F1 than hard‑filter baselines by isolating compact, high‑precision regions of $p_C$ even if they are embedded within broader regions exhibiting some noise ($p_N$).

\subsection{Implementation and Noise Handling}
Implemented using Python (XGBoost \cite{chen2016xgboost}, scikit-learn, SHAP). LLM feature synthesis employed meta-llama/Llama-4-Maverick-17B-128E-Instruct-FP8 with deterministic settings (temperature=0). Key parameters ($\alpha$ range [2, 10], tree depth max 5) are configurable. Robustness to noise stems from explicit noise modeling via the dedicated noise prediction model, weight-based soft exclusion (Eq. \ref{eq:weight_def_alt}), and adaptive $\alpha$ optimization. 
\section{Experimental Methodology and Evaluation Metrics}
\label{sec:exp_meth}

We evaluate \textsc{SIFOTL} in two complementary regimes:
\emph{(i)} a \textbf{public, semi-synthetic MEPS benchmark} that embeds a realistic policy intervention in noisy survey data, and
\emph{(ii)} the \textbf{synthetic stress-suite} covering a broad grid of shift and noise types.
All modeling is performed on privacy-compliant summary tables; no raw rows leave the local system.

\vspace{-0.75\baselineskip}
\subsection{MEPS test dataset}
\label{ssec:meps}

\paragraph{Data.}
We use Round 1 of the Medical Expenditure Panel Survey (MEPS) Panel 24 public-use file for participants over age 55 ($N{=}2\,077$) \cite{meps2024}. The rule-based noise analysis module labels 105 rows as \texttt{is\_noisy}=1.

\paragraph{Intervention.}
A respondent is \emph{policy-eligible} if $\text{Age} \ge 55$, enrolled in Medicare, diagnosed with diabetes, and income below 200\% of the federal poverty level, and has had positive self/family funded prescription spend.
 There are $167$ such rows in the control table. In a cloned test table, we set the prescription spend \texttt{RXSLFY}$:=0$ for each eligible record with probability $0.9$ if the \texttt{is\_noisy} flag is not set and with probability $0.3$ if the \texttt{is\_noisy} flag is set, 
 producing 148 shifted rows.

\vspace{-0.5\baselineskip}
\subsection{Synthetic stress-suite}
\label{ssec:synthetic_suite}

To test edge cases absent from MEPS, we retain the
eighteen synthetic scenarios based on Synthea's ABM simulation.
We use users' medical records from two years.
We've shared Synthea's data generation settings in our GitHub repository.

\vspace{-0.5\baselineskip}
\subsection{Ground-truth interventions}
\label{ssec:interventions} We clone each year's dataset with the following interventions.
\begin{itemize}[leftmargin=*]
  \item \textbf{T1 – Cost uplift (+20\,\%).}\;
        \texttt{TOTAL\_CLAIM\_COST} multiplied by $1.2$
        for rows with \texttt{TOT\_INCOME}$\ge150\,000$,
        \texttt{AGE}$>59$, \texttt{TOTSLFY}$\ge100\,000$
        and \texttt{PAYER\_NAME}=\textsc{Medicare}.
  \item \textbf{T2 – Coverage reduction (–30\,\%).}\;
        \texttt{PAYER\_COVERAGE} scaled by $0.7$
        for male patients in six Massachusetts counties
        whose encounters are labeled ambulatory, wellness, or home.
  \item \textbf{T3 – Encounter-cost jitter.}\;
        Add $\mathcal{N}(0,30)$ to \texttt{BASE\_COST}
        for divorced men aged over 40.
\end{itemize}

\vspace{-0.5\baselineskip}
\subsection{Observational noise}
\label{ssec:noise}

Further, we generate two noise regimes for every cloned table with the above interventions.
\textbf{N1} randomly injects a targeted 5\% to 10\%  noise rate; \textbf{N2} injects a targeted 10\% to 15\% per mechanism.
Mechanisms for noise in T1 intervention were duplicate rows and outliers (TOTAL\_CLAIM\_COST set to 3x to 5x value) based on values of ENCOUNTERCLASS and REASONDESCRIPTION fields (selected because with high cardinality of these text fields, noisy changes could be made without generating a coherent counter-signal). For the T2, noise mechanisms were missing values in PAYER\_COVERAGE and rounding of the values based on payer, gender, and marital status. For the T3, set BASE\_COST to zero and REASONDESCRIPTION corruption based on the encounter class and the reason code. In each case, the noise row selection criteria were independent from the intervention selection criteria, with overlapping changes. Each intervention was applied to each year's simulation-generated table, followed by two noise regimes per such clean intervention table, yielding a total of 18 test datasets for our testing.

\subsection{Research questions and metrics}
\label{ssec:rqs}
\begin{enumerate}[label=\textbf{RQ\arabic*:\ }, leftmargin=*]
  \item \emph{Segment accuracy.} Can a method recover the ground-truth slice under noise?
  \item \emph{Feature-synthesis benefit.} Does LLM-guided feature engineering improve Stage-1 XGBoost?
  \item \emph{Robustness. How do \textsc{SIFOTL} 's internal metrics respond to different interventions and noise levels?}
\end{enumerate}
\vspace{-0.5\baselineskip}
\subsection{Baselines}
\label{ssec:data_baselines}
To compare our results with existing methods, we use two baselines:
(i) Google BigQuery Contribution Analysis Model (BQCA) \cite{google2023bqca}, which uses a combination of SQL queries and machine learning to identify the contribution of different features to the target variable. It is a black-box model but provides a good baseline as a commercial solution accessible to most EHR teams.
(ii) a $\chi^2$ / Point-Biserial correlation screen with FDR correction as a statistical baseline. This approach is widely used in the literature for identifying significant features in high-dimensional data. It is a simple and interpretable method that can be easily implemented, making it a good baseline for comparison with more complex models.
For both these baselines, we used standard configuration options and hyperparameters (available in our GitHub repository)

\section{Results and Analysis}

\subsection{Overall Performance Comparison (RQ1)}
\label{ssec:overall}

\paragraph{Real survey.}
For the MEPS test described in \S\ref{ssec:meps}, against a ground-truth slice of 148 subsidised respondents, \textsc{SIFOTL}
recovers the segment with \textbf{F1\,=\,0.85}; BigQuery Contribution Analysis
drops to 0.46 and the statistical tests to 0.20.

\paragraph{Simulation data stress-suite.}
Table~\ref{tab:cross_model_f1}  summarises performance across the
18 synthetic scenarios. \textsc{SIFOTL} again achieves the highest scores, sustaining
0.86–0.96 F1 under \texttt{N0} and remaining above 0.75 even when
observational noise is injected, whereas baseline scores span 0.19–0.67.
The pattern mirrors the MEPS result: high precision from explicit noise
handling, and there is a high recall from the twin-model architecture.

\begin{table}[h]
\centering
\caption{Cross-Model Performance Comparison: F1 Score for Intervention Segment Identification (Avg. over two control/test table pairs)}
\label{tab:cross_model_f1}
\begin{tabular}{l c c c c}
\toprule
& Noise Level & F1 SIFOTL & F1 BQCA & F1 StatsTest \\
\midrule
\multirow{3}{*}{T1} & N0 & \textbf{0.955} & 0.671 & 0.414 \\
  & N1 & \textbf{0.871} & 0.245 & 0.364 \\
  & N2 & \textbf{0.872} & 0.261 & 0.355 \\
\midrule
\multirow{3}{*}{T2} & N0 & \textbf{0.862} & 0.373 & 0.556 \\
  & N1 & \textbf{0.757} & 0.335 & 0.565 \\
  & N2 & \textbf{0.769} & 0.357 & 0.564 \\
\midrule
\multirow{3}{*}{T3} & N0 & \textbf{0.908} & 0.414 & 0.455 \\
  & N1 & \textbf{0.889} & 0.192 & 0.455 \\
  & N2 & \textbf{0.882} & 0.192 & 0.455 \\
\bottomrule
\end{tabular}
\end{table}

\subsection{Feature Generation Effectiveness (RQ2)}
\label{ssec:feature_eff}

LLM-guided feature synthesis enhances Stage-1 (XGBoost) classifier accuracy, achieving improvements of up to 0.3 percentage points in the simulation data suite with Featuretools{\cite{7344858} based features. It also yields a  0.24 percentage point accuracy gain on the MEPS task. Although the absolute lift is small, it corresponds to a relative reduction in residual errors over an already strong XGBoost baseline with features generated from FeatureTools library. While these observed gains vary by specific scenario and noise level and were not calibrated for generalizability across diverse datasets, the benefit from LLM features in creating interpretable segments based on data interaction features for most test scenarios (features data is available in our shared Github repository) highlights the potential benefits of LLM-based feature engineering in both simulation data and real-world survey settings.

\subsection{Robustness and Noise Handling (RQ3)}
SIFOTL's components demonstrated robust performance. Stage 1 classifiers (Table \ref{tab:sifotl_stage1}) maintained high accuracy ($>93\%$) even under significant noise. The Stage 2 decision tree (Table \ref{tab:sifotl_stage2}) adaptively balances signal coverage ($M_{signal}$) with noise robustness ($M_{noise}$) through Pareto optimization of the weighting penalty $\alpha$. Across all noise levels, $M_{signal}$ remained high, indicating effective identification of intervention-affected segments. The achieved $M_{noise}$ was generally moderate, reflecting the inherent difficulty of perfectly separating complex signals from certain types of noise. For instance, the positive leaf-level correlations between $p_C$ and $p_N$ (leading to lower $M_{noise}$ for T1) can occur when intervention and specific noise mechanisms impact similar features. SIFOTL's soft-weighting strategy (Eq. \ref{eq:weight_def_alt}) is designed for such scenarios; by discounting rather than eliminating high-$p_N$ instances, it allows the decision tree to identify high-purity $p_C$ "pockets" even when embedded in noisy regions. This contributes to the strong final segment F1-scores (Table \ref{tab:cross_model_f1}) despite these internal correlations. Finally, the positive correlation $\rho(p_C, \text{mask})$ in Table \ref{tab:sifotl_stage2} confirmed that the extracted segments effectively captured instances with high intervention probability.

\begin{table}[h]
\centering
\caption{SIFOTL Stage 1 (Intervention Classifier) Performance Summary (Avg. over two control/test table pairs)}
\label{tab:sifotl_stage1}
\begin{tabular}{l c c c}
\toprule
 & Noise Level & XGB Accuracy (\%) & Top Feat. |SHAP| (log-odds) \\
\midrule
\multirow{3}{*}{T1} & N0 & 99.85 & 5.177 \\
  & N1 & 93.45 & 1.992 \\
  & N2 & 94.94 & 2.154 \\
\midrule
\multirow{3}{*}{T2} & N0 & 98.85 & 2.946 \\
  & N1 & 93.13 & 2.027 \\
  & N2 & 95.15 & 1.558 \\
\midrule
\multirow{3}{*}{T3} & N0 & 99.99 & 5.416 \\
  & N1 & 97.96 & 3.105 \\
  & N2 & 98.92 & 2.706 \\
\bottomrule
\end{tabular}
\\ \vspace{1mm}
\end{table}

\begin{table}[h]
  \centering
  \caption{SIFOTL Stage 2 (Decision Tree) Performance Summary (Noisy Conditions, Avg. over two control/test table pairs)}
  \label{tab:sifotl_stage2}
  \footnotesize
  \begin{tabular}{l c c c c c}
    \toprule
     & Noise Level & Optimal avg. $\alpha$ & $M_{signal}$ & $M_{noise}$ & $\rho(p_C, \text{mask})$ \\
    \midrule
  \multirow{2}{*}{T1} & N1 & 7.1 & 942 $\pm$ 157 & 0.364 & 0.64 \\
  & N2 & 6.4 & 939 $\pm$ 215 & 0.428 & 0.57 \\
\midrule
\multirow{2}{*}{T2} & N1 & 6.9 & 1,469 $\pm$ 465 & 0.915 & 0.08 \\
  & N2 & 6.2 & 1,174 $\pm$ 95 & 0.751 & 0.25 \\
\midrule
\multirow{2}{*}{T3} & N1 & 6.0 & 492 $\pm$ 249 & 0.684 & 0.32 \\
  & N2 & 5.9 & 462 $\pm$ 262 & 0.510 & 0.49 \\
    \bottomrule
  \end{tabular}
 
\end{table}

\subsection{Qualitative Analysis and Interpretability}
Beyond quantitative metrics, SIFOTL demonstrated strong qualitative advantages. While the broader concept of model interpretability has its complexities \cite{lipton2018mythos}, the decision tree rules generated in Stage 2 provided clear, human-understandable characterizations of the identified segments. For instance, for the T1 Cost Uplift intervention, SIFOTL consistently produced rules that closely matched the definition of ground truth. In contrast, BQCA often produced fragmented rules, hindering interpretability, and suffered in handling numeric fields such as Age. Statistical tests typically identified overly broad segments or missed key interaction effects. The noise-aware optimization was crucial; analysis showed SIFOTL segments contained significantly fewer records flagged by the noise inference step compared to baseline segments. Thus, our experiments confirm that SIFOTL advances robust data shift driver detection by: (1) outperforming baselines, especially under noise, with high F1 scores. (2) Benefiting from privacy-preserving LLM feature synthesis based on statistical summaries. (3) Maintaining robustness across diverse scenarios via its noise-aware twin-model architecture and adaptive optimization. (4) Effectively disentangling signal from noise, yielding cleaner, more interpretable segments.

\section{Related Work}
SIFOTL draws upon and extends several research areas focused on data shift detection, noise robustness, and privacy-preserving analytics.

\paragraph{Statistical Tests and Analysis.} Traditional statistical methods for detecting distribution shifts provide foundational frameworks but often lack mechanisms for multivariate segment identification or explicit noise handling. Key approaches include Kolmogorov-Smirnov tests, Chi-squared tests, Maximum Mean Discrepancy \cite{gretton2012kernel}, and Wasserstein distance \cite{Panaretos_2019}. While these methods establish statistical frameworks for identifying when distributions differ, they typically struggle with pinpointing specific segments driving the change and handling noise simultaneously.

\paragraph{Drift Detection in ML.} Dataset shift detection and adaptation have been extensively studied in the machine learning literature. Approaches that frame drift detection as a classification problem \cite{gama2014survey} or monitor feature attribution changes \cite{lundberg2017unified} have shown promise. However, as demonstrated by Rabanser et al. \cite{rabanser2019failing}, many drift detection methods fail under complex, real-world conditions with noise—a limitation SIFOTL specifically addresses through its noise-robust architecture. The challenge of concept drift adaptation \cite{bifet2009adaptive} becomes particularly important in healthcare applications \cite{finlayson2021clinician,subbaswamy2020development}, where Saria and Subbaswamy \cite{saria2019tutorial} highlight the importance of developing shift-stable models in evolving clinical contexts. Multiple testing correction methods like Benjamini-Hochberg \cite{benjamini1995controlling} help control false discoveries in high-dimensional analyses but lack mechanisms for coherent segment identification.

\paragraph{LLM-Assisted Feature Generation.} Recent developments in feature engineering using large language models show promise in tabular data analysis. While approaches using LLMs for tabular data \cite{hollmann2023large} \cite{nam2024optimizedfeaturegenerationtabular} \cite{balek2024llmbasedfeaturegenerationtext}demonstrate potential for incorporating domain knowledge, SIFOTL differentiates itself by using a structured workflow with LLMs that use only statistical summaries of data and preserve data privacy and doesn't expose sensitive data. 

\paragraph{Privacy-Safe Data Handling.} The increasing restrictions on data access, particularly in healthcare under regulations like HIPAA, significantly impact analytical capabilities \cite{ness2007influence,wartenberg2010privacy}. Rodriguez \cite{rodriguez2024native} documents how these limitations severely impact Native American public health officials, creating "blind spots" that impede critical analyses. While synthetic data generation approaches \cite{walonoski2018synthea} can help with model validation, they don't address the core challenge of noise separation while preserving privacy. Commercial contribution analysis tools \cite{google2023bqca} identify influential segments but generally lack SIFOTL's specific noise-handling capabilities while maintaining privacy compliance.

SIFOTL advances the knowledge in these areas by integrating such research work into a cohesive framework that simultaneously addresses segment identification, noise robustness, and privacy preservation.

\section{Conclusion}
This paper introduces SIFOTL, a principled approach for identifying population segments driving data shifts while effectively handling observational noise and privacy constraints. Our twin-model architecture, privacy-preserving LLM-guided feature synthesis, and Pareto-weighted decision tree optimization substantially improve existing methods. Across real-world survey data and diverse synthetic scenarios, SIFOTL consistently achieves superior F1 scores compared to baseline approaches, even under challenging noise conditions. The method produces interpretable segment definitions that enable actionable insights while maintaining privacy compliance, which we expect to help the analysts and public policy experts in their data-driven decision-making.

\subsection{Limitations and Future Work}

Several limitations guide our future research. While our evaluation used interventions in real-world scenarios, they could not be expected to capture all the complexities fully, and our simulated noise represents only a subset of potential data quality issues. The method's performance depends partly on the LLM component, which requires offline cleanup for generated code. Formal privacy guarantees (e.g., differential privacy) remain unestablished for our statistical summary generation. Additionally, the twin-model architecture adds computational overhead that could benefit from optimization for larger datasets. Future work should address these limitations through real-world validation, alternative feature synthesis methods, computational optimizations, formalized privacy guarantees, causal inference integration, and multi-modal data handling. Despite these limitations, SIFOTL contributes to the research community through its structured, noise-robust, privacy-conscious approach to data shift detection.

\balance
\newpage
\bibliographystyle{ACM-Reference-Format}
\bibliography{references}

\newpage
\appendix

\section{Reproducibility}

To facilitate reproducibility and further research, we provide access to the following dataset and source code files:

\subsection{Datasets}

\begin{itemize}
    \item \textbf{Baseline Control Datasets:}
    We include the link to download the MEPS HC-245 PANEL 24 dataset and provide CSV files of simulated EHR data from Synthea\cite{walonoski2018synthea}.
    \item \textbf{Clean Intervention Datasets:} Six datasets with ground truth intervention flags.
    \item \textbf{Noisy Intervention Datasets:} Twelve datasets with injected noise at two levels, including ground truth noise labels along with the MEPS HC-245 panel dataset with intervention
    \item \textbf{Noise Inference Rules:} We share the specific data quality checklist and noisy label inference rules used in the experimental setup. The noise model is trained on these inferred labels and has no visibility to the ground truth of the noise
\end{itemize}

\subsection{Statistical Test Results}

\begin{itemize}
    \item Complete outputs from the baseline statistical tests ($\chi^2$ test, Point-Biserial correlation) for all 18 scenarios and the MEPS HC-245 dataset.
    \item FDR-corrected q-values and effect sizes used for the statistical tests baseline's segment identification.
    \item Detailed statistical summaries provided to the LLM for feature synthesis, including distributions, correlations, and significance tests comparing relevant groups, including details about the suppressed slices for privacy reasons.
\end{itemize}

\subsection{Training Results}

\begin{itemize}
    \item XGBoost model hyperparameters for both the intervention and noise prediction models across all scenarios.
    \item Feature importance rankings and SHAP values from all trained models.
    \item Decision tree parameters and resulting tree structures from Stage 2.
    \item Pareto frontier analysis results, including tested $\alpha$ values and corresponding metrics ($M_{signal}$, $M_{noise}$).
\end{itemize}

\subsection{Baseline Configuration Settings}
\begin{itemize}
\item Settings used for BigQuery analysis and statistical tests baseline
\end{itemize}

\subsection{Evaluation Metrics}

\begin{itemize}
    \item Complete F1 scores, precision, and recall for segment identification across all methods and scenarios (as summarized in Table \ref{tab:cross_model_f1}).
    \item Internal model performance metrics (accuracy, signal coverage, noise robustness) as shown in Tables \ref{tab:sifotl_stage1} and \ref{tab:sifotl_stage2}.
   
\end{itemize}

\subsection{Source Code and LLM Interaction}

\begin{itemize}
    \item The configuration, twin xgboost, and segment builder files enable running segment results on every test dataset. The configuration file is flexible to run similar tests on any additional datasets  
    \item We also include the prompt templates used for feature synthesis, which can be used if the new datasets are to be tested (for current datasets, the LLM-generated features are already in the datasets). 
    \item LLM-generated feature definitions (step 1 of interaction) and corresponding Python implementation code (before manual cleaning/validation when it failed in limited cases) (step 2) are also included for reference purposes.
    \item We use LLM Model meta-llama/Llama-4-Maverick-17B-128E-Instruct-FP8 (latest available during our experiments) with temperature=0 for reproducibility.
\end{itemize}

You can access these files through our project website that links to a GitHub repository at \url{https://datascience.healthcare/sifotl-github}. We hope this comprehensive set of resources will enable researchers to fully validate our results and build upon SIFOTL for future advancements in robust data shift detection.

\end{document}